\documentclass[11pt]{article}

\usepackage[preprint]{acl}

\usepackage{times}
\usepackage{latexsym}

\usepackage[T1]{fontenc}

\usepackage[utf8]{inputenc}

\usepackage{microtype}

\usepackage{inconsolata}

\usepackage{graphicx}
\usepackage{algorithm}
\usepackage{amsmath}
\usepackage{amssymb}
\usepackage{makecell}
\usepackage{booktabs}       
\usepackage{amsfonts}       
\usepackage{nicefrac}       
\usepackage{xcolor}         
\usepackage{varwidth}
\usepackage{subcaption}
\usepackage{adjustbox}
\usepackage{colortbl}
\usepackage{multirow}
\usepackage{multicol}
\usepackage{algpseudocode}
\usepackage{amsthm}
\title{CompassOPD: Cross-Family On-Policy Distillation via Within-Family Likelihood Shifts}

\author{Naibin Gu\textsuperscript{\rm 1,2},\ Qingyi Si\textsuperscript{\rm 3},\ Chenxu Yang\textsuperscript{\rm 1,2},\ Chuanyu Qin\textsuperscript{\rm 1,2},\ Junhao Zhou\textsuperscript{\rm 1,2},\\{\bf Peng Fu\textsuperscript{\rm 1,2}\thanks{\ \ Corresponding author.},\ Zheng Lin\textsuperscript{\rm 1,2},\ Weiping Wang\textsuperscript{\rm 1}} \\ 
\textsuperscript{\rm 1}Institute of Information Engineering, Chinese Academy of Sciences, Beijing, China \\
\textsuperscript{\rm 2}School of Cyber Security, University of Chinese Academy of Sciences, Beijing, China \\
\textsuperscript{\rm 3}JD.COM \\
  \texttt{ \textrm{\{}gunaibin,fupeng\textrm{\}}@iie.ac.cn} \\
}

\begin{document}
\maketitle
\begin{abstract}
On-policy distillation (OPD) provides dense token-level supervision on student-generated trajectories. Although OPD performs strongly when teacher and student belong to the same model family, we find that its effectiveness degrades in cross-family settings even after tokenizer alignment, with substantially stronger external teachers offering little additional improvement. To understand this disconnect, we decompose the cross-family OPD signal into two components: an offset between a low-capability teacher-family reference and the student, and the within-family log-likelihood shift from that reference to the strong teacher. Standard OPD transfers both components together, allowing the offset to dominate the update direction and obscure the changes associated with teacher capability improvements. We propose CompassOPD, which removes this offset and transfers the within-family shift, while a frozen student reference anchors updates to the student's initial policy. Thus, both teacher-side and student-side changes are measured within their respective model families. Experiments across three student families and multiple teacher families show that CompassOPD consistently outperforms standard cross-family OPD, improving average reasoning accuracy by up to 5.50 points. For an MoE teacher, we further construct the reference directly from the teacher checkpoint by reducing expert activation, eliminating the need for a separate reference checkpoint while retaining a 3.43-point gain over OPD.
\end{abstract}

\section{Introduction}
On-policy distillation (OPD) has emerged as an effective paradigm for post-training large language models~\cite{agarwal2024onpolicydistillationlanguagemodels,gu2026minillmonpolicydistillationlarge,lu2025onpolicydistillation}. It samples trajectories from the student policy and uses the teacher's token-level likelihoods to provide dense supervision along the states that the student actually visits. This on-policy interaction supplies token-level learning signals while avoiding the distribution gap introduced by teacher-generated training data. By combining on-policy exploration with dense token-level feedback, OPD has shown strong empirical gains on reasoning tasks~\cite{yang2026selfdistilledrlvr,yang2026learningteachergeneralizedonpolicy,gu2026coevolvingpolicydistillation}.

However, existing OPD methods have mainly been studied in scenarios where the teacher and student models belong to the same model family.\footnote{We use ``model family'' to denote a release series, such as Qwen3,
Qwen3.5, or Ministral-3. Different families often differ in vocabulary
and training recipes.} Such models largely share training recipes and policy structures, with capability differences typically arising from model scale or post-training strategies~\cite{li2026rethinkingonpolicydistillationlarge,ma2026mopdmultiteacheronpolicydistillation}. In practice, the student's model family may not provide a sufficiently strong teacher model, making it necessary to use stronger models from other families. Prior work has addressed the representational differences caused by different tokenizers through vocabulary or text-space alignment, enabling cross-model likelihood comparison. Such alignment makes likelihood scores comparable over shared textual units~\cite{sun2026simctrecoveringlostsupervision,niu2026breakingtokenizerbarrieronpolicy,wang2026crosstokenizeronpolicydistillationbyteprefix}, but leaves open whether the resulting supervision effectively transfers teacher capabilities across model families.

\begin{figure*}[t]
    \centering

    \subfloat[Same-family vs.\ cross-family OPD.]{
        \includegraphics[width=0.26\textwidth]
        {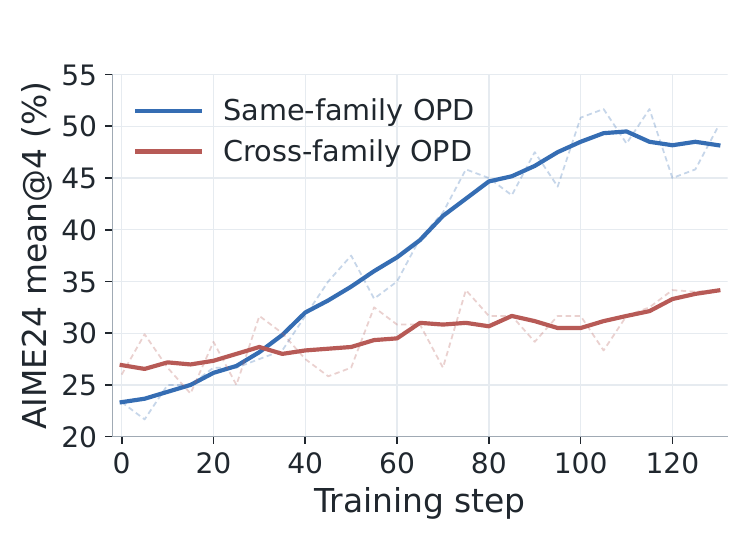}
        \label{fig:motivation_cross_family}
    }
    \hfill
    \subfloat[Transfer of the teacher capability gap.]{
        \includegraphics[width=0.30\textwidth]
        {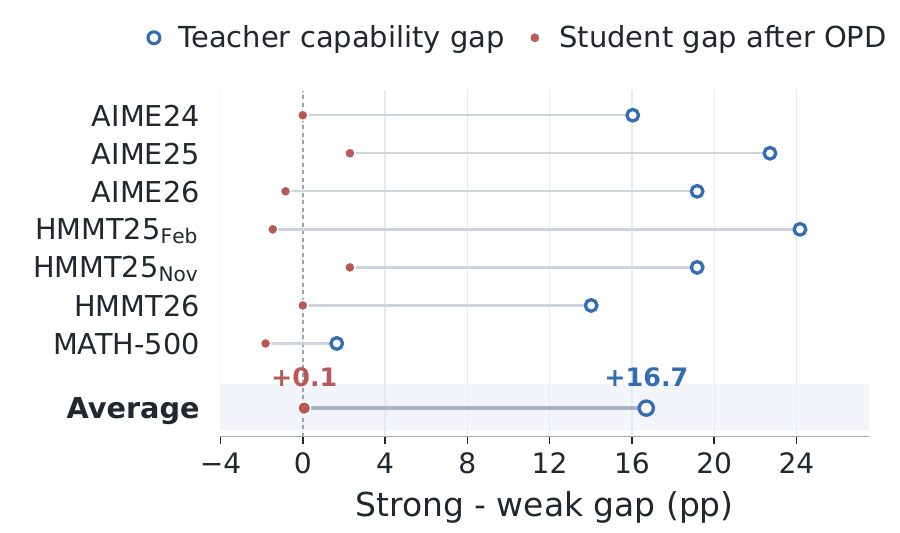}
        \label{fig:motivation_teacher_gap}
    }
    \hfill
    \subfloat[OPD signal similarity within the teacher family.]{
        \includegraphics[width=0.38\textwidth]
        {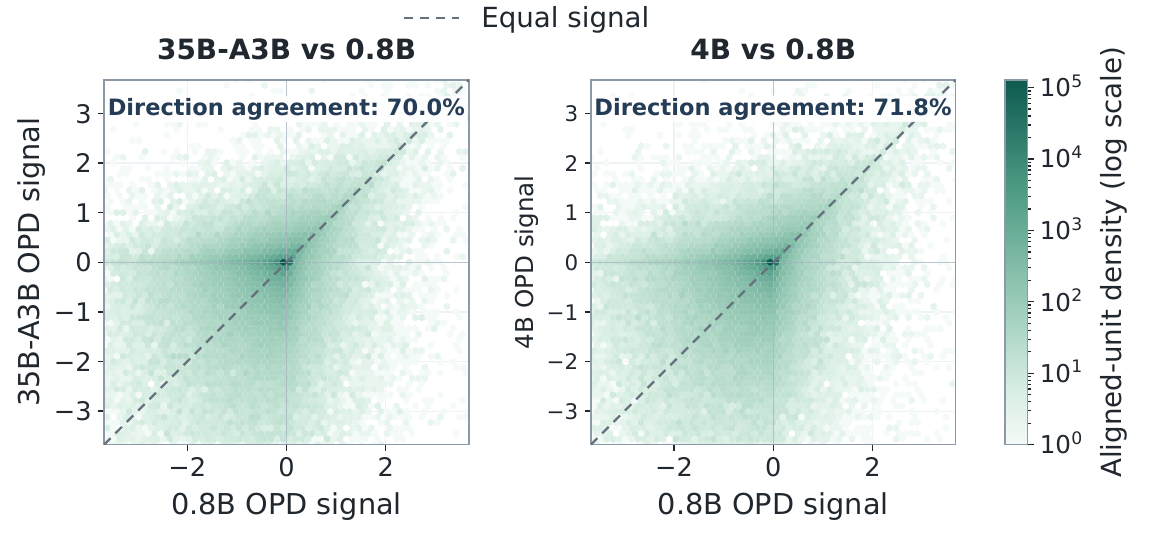}
        \label{fig:motivation_signal}
    }

    \caption{
    Motivating observations for cross-family OPD.
    (a) Performance comparison of same-family and cross-family OPD. The same-family teacher model uses Qwen3-30B-A3B-Instruct, while the cross-family teacher model uses Qwen3.5-35B-A3B; the student model is Qwen3-4B in both cases.
    (b) The performance gap between Qwen3.5-35B-A3B and Qwen3.5-4B, and the gap between the Qwen3-4B student models distilled from these two teacher models. Hollow and solid markers denote the teacher model gap and the distilled student model gap, respectively.
    (c) When scoring the same aligned student-generated actions, Qwen3.5-0.8B agrees in update direction with Qwen3.5-35B-A3B and Qwen3.5-4B at approximately $70\%$ of positions, revealing a substantial OPD component shared across teacher scales.
    }
    \label{fig:motivation}
\end{figure*}

We find that, despite vocabulary alignment, replacing a same-family teacher model with a cross-family teacher model of similar capability does not preserve the benefits of OPD. Figure~\ref{fig:motivation}(a) shows that same-family OPD produces sustained improvements, whereas cross-family OPD plateaus quickly. Moreover, as shown in Figure~\ref{fig:motivation}(b), external teacher models with substantially different capabilities yield distilled student models with nearly identical performance. Thus, under cross-family OPD, improvements in teacher capability do not reliably translate into greater gains in student performance.

To understand this phenomenon, we further compare the OPD update directions produced by models of different scales within the same teacher family on identical student trajectories. Surprisingly, as shown in Figure 1(c), despite the substantial capability gap, the strongest teacher and the smallest model in the teacher family still assign the same update direction on 70\% of aligned textual units. This prompts us to ask: which part of the strong teacher signal truly reflects its capability improvement? Using a low-capability model from the same family as a reference, we find that the cross-family OPD signal can be decomposed into two components: one is the pre-existing cross-family offset between the reference model and the student, and the other is the within-family likelihood change from the low-capability reference model to the strong teacher model. Standard OPD transfers both components together; when the cross-family offset determines the sign of the combined signal, the update follows the direction already induced by the low-capability reference, while the strong teacher's additional contribution changes only its magnitude.

Based on this observation, we propose \textbf{CompassOPD}. CompassOPD removes the cross-family offset and transfers only the within-family likelihood change. A frozen copy of the student's initial policy provides an anchor that modulates the update according to the student's displacement from that initialization. In this way, both teacher-side and student-side variations are measured within their respective model families. Through this design, CompassOPD transfers the policy change directions associated with teacher capability improvement rather than directly matching the absolute likelihoods of a cross-family teacher model.

Experiments across three student families and multiple teacher families show that CompassOPD consistently outperforms standard cross-family OPD under matched student initializations, improving average reasoning accuracy by up to 5.50 points. Further analysis shows that restoring the removed cross-family offset degrades performance. With a fixed teacher-family reference, CompassOPD also yields progressively better student performance as teacher scale increases. For MoE teacher models, we further construct a low-capability reference model from the teacher checkpoint itself by reducing expert activation, thereby eliminating the need for a separately released reference model while still retaining substantial improvements over OPD.
\section{Preliminaries}
\label{sec:preliminaries}
\paragraph{On-Policy Distillation.}
Given a prompt $x$, the student policy $\pi_\theta$ generates a response $y \sim \pi_\theta(\cdot \mid x)$. At generation step $t$, let $c_t=(x,y_{<t})$ denote the student-visited context and let $y_t$ be the sampled action. OPD~\cite{lu2025onpolicydistillation} asks a teacher $\mathcal{T}$ to score the same action under the same textual context and constructs the token-level signal
\begin{equation}
A_t^{\mathrm{OPD}}
=
\log \mathcal{T}(y_t\mid c_t)
-
\log \pi_\theta(y_t\mid c_t).
\label{eq:sampled_opd_signal}
\end{equation}
The detached signal is used as an advantage in the policy update, , with positive values encouraging the sampled action and negative values discouraging it:
\begin{equation}
\mathcal{L}_{\mathrm{OPD}}
=
-
\mathbb{E}_{y\sim\pi_\theta(\cdot\mid x)}
\left[
A_t^{\mathrm{OPD}}
\log \pi_\theta(y_t\mid c_t)
\right],
\label{eq:sampled_opd_loss}
\end{equation}
Because the teacher evaluates trajectories generated by the current student, OPD provides dense supervision on states that the student actually visits.
\paragraph{Cross-Tokenizer Likelihood Alignment.}
Equation~\ref{eq:sampled_opd_signal} assumes that the teacher and student represent the sampled action with compatible tokens. Prior work on cross-tokenizer distillation addresses this mismatch by aligning model likelihoods in text space~\cite{sun2026simctrecoveringlostsupervision,niu2026breakingtokenizerbarrieronpolicy}. Following this line of work, we decode the student response into text, re-encode it with the teacher tokenizer, and align the two token
sequences according to their shared textual boundaries. This procedure partitions the response into aligned textual units
\begin{equation}
\mathcal{U}(y)
=
\{z_1,z_2,\ldots,z_M\},
\label{eq:aligned_units}
\end{equation}
where each unit is either a one-to-one token match or a span represented by different numbers of teacher and student tokens. We use $\ell_{\mathcal{M}}(z_u\mid c_u)$ to denote the aligned log-likelihood score assigned by model $\mathcal{M}$ to unit $z_u$ under its preceding textual context $c_u$. When considering an individual aligned unit, we omit the index $u$ and write $(z,c)$. We apply the same text-space alignment procedure to standard cross-family OPD and CompassOPD, ensuring that differences in the alignment procedure do not confound their comparison.
\section{Understanding Cross-Family OPD}
\label{sec:pilot}
When the teacher model and the student model come from the same model family, for example both being Qwen3 series models, OPD has already demonstrated significant performance gains~\cite{yang2025qwen3technicalreport}. A natural question is whether these gains still hold when the teacher model is switched to a model from a different family.

\subsection{Same-Family and Cross-Family OPD}
We first consider a relatively close model family for comparison. In the same-family setting, we distill from a Qwen3-30B-A3B-Instruct teacher model to a Qwen3-4B student model. In the cross-family setting, we use Qwen3.5-35B-A3B~\cite{qwen3.5} as the teacher model. To reduce differences in output style, the cross-family student model is first initialized by performing SFT on trajectories generated by the teacher (see Appendix~\ref{app:implementation} for details).

The experimental results are shown in Figure~\ref{fig:motivation}(a). Same-family OPD produces sustained improvements in reasoning accuracy, whereas cross-family OPD plateaus after limited early gains despite the preceding teacher-trajectory SFT.

\paragraph{Is the performance limitation of cross-family OPD related to teacher capability?} After observing the limited gains achieved with Qwen3.5-35B-A3B, we examine whether student gains track teacher capability. If a weaker cross-family teacher were used, the gains from distillation would be even worse, while a stronger teacher should correspondingly improve performance. We compared two teachers from the Qwen3.5 family but with significantly different performance levels, Qwen3.5-4B and Qwen3.5-35B-A3B, for performing OPD on the same Qwen3-4B student model.

Figure~\ref{fig:motivation}(b) reveals a clear misalignment between teacher capability and student model improvement. Across benchmarks, Qwen3.5-35B-A3B outperforms Qwen3.5-4B by 16.7 points on average. However, the Qwen3-4B student models distilled from these two teachers differ by only 0.1 points. On several benchmarks, the student distilled from the stronger teacher performs no better, and sometimes slightly worse, than the student distilled from the weaker teacher. Thus, the large capability gap within the teacher family translates into only marginal differences in the gains achieved by cross-family OPD.

Taken together, these observations lead to a more specific question: \textit{Why does the large capability gap between the two teachers translate into only a marginal difference in student model gains under cross-family OPD?}
\begin{figure*}[t]
  \centering
  \includegraphics[width=0.9\textwidth]{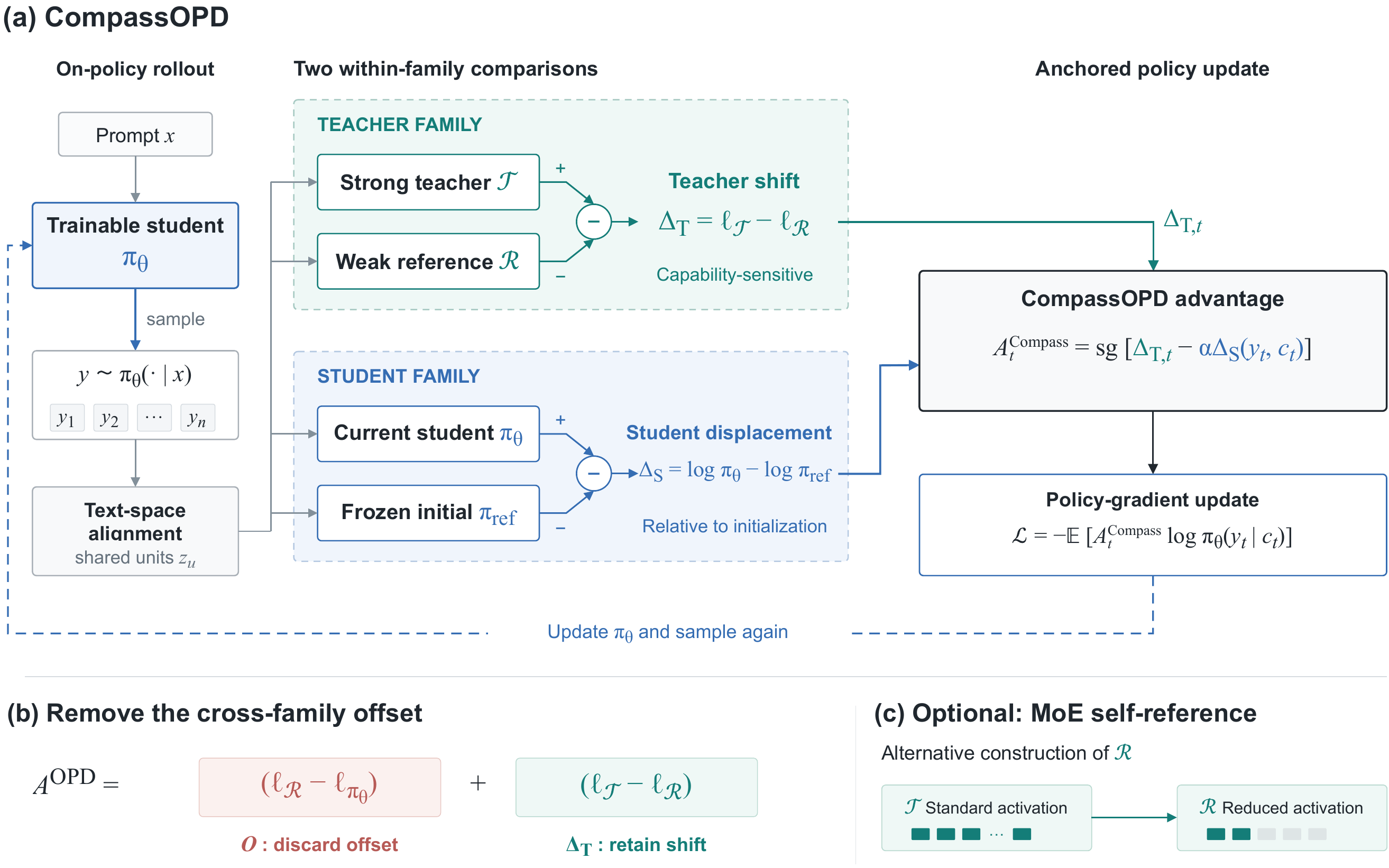}
  \caption{
  Overview of CompassOPD. Given student-generated trajectories, the
  strong teacher $T$ and its within-family reference $R_T$ score the
  same aligned textual actions. Their likelihood difference isolates
  the capability-sensitive within-family shift by removing the
  reference-anchored cross-family offset. The resulting signal is
  combined with the student's displacement from its initial policy
  to guide the on-policy update.
  }
  \label{fig:compassopd-method}
  \end{figure*}
\subsection{Dissecting cross-family OPD signals}
To understand why the large teacher capability gap in Figure~\ref{fig:motivation}(b) leads to only marginal differences in student performance, we examine how token-level OPD supervision signals vary with teacher capability.
We use Qwen3.5-0.8B, Qwen3.5-4B, and Qwen3.5-35B-A3B to score the same cross-family student trajectories, with these three models covering a wide range of reasoning performance within the same model family.

For an aligned textual unit $z$ under context $c$, the OPD signal produced by a scoring model $\mathcal{M}$ is
\begin{equation}
A_{\mathcal{M}}(z,c)
=
\ell_{\mathcal{M}}(z\mid c)
-
\ell_{\pi_\theta}(z\mid c).
\label{eq:pilot_opd_signal}
\end{equation}
Its sign determines whether OPD increases or decreases the likelihood of that observed action.

Figure~\ref{fig:motivation}(c) compares the signals produced by Qwen3.5-0.8B with those produced by the two larger models.
Surprisingly, Qwen3.5-0.8B assigns the same update direction as Qwen3.5-35B-A3B on $70.0\%$ of the aligned units, and the same direction as Qwen3.5-4B on $71.8\%$ of the aligned units.
Despite the large capability differences, the two larger models frequently induce the same update direction as the low-capability model, even though the latter has weaker reasoning abilities than the cross-family student.
This motivates using the low-capability model as a reference to separate the OPD signal it already induces from the likelihood changes introduced by a stronger teacher.

\paragraph{What masks the updates introduced by cross-family teachers?}
For the same student action $z$ under context $c$, let $\mathcal{T}$ denote the strong teacher and $\mathcal{R}$ a low-capability reference from the same teacher family.
Using $\mathcal{R}$ as the OPD teacher yields:
\begin{equation}
A_{\mathcal{R}}(z,c)
=
\ell_{\mathcal{R}}(z\mid c)
-
\ell_{\pi_\theta}(z\mid c).
\label{eq:reference_opd_signal}
\end{equation}
For the strong teacher $\mathcal{T}$, adding and subtracting the reference score gives:
\begin{equation}
\begin{aligned}
&A_{\mathcal{T}}(z,c)
=
\ell_{\mathcal{T}}(z\mid c)
-
\ell_{\pi_\theta}(z\mid c) \\
=
&\underbrace{
\ell_{\mathcal{R}}(z\mid c)
-
\ell_{\pi_\theta}(z\mid c)
}_{O(z,c)} 
+
\underbrace{
\ell_{\mathcal{T}}(z\mid c)
-
\ell_{\mathcal{R}}(z\mid c)
}_{\Delta_T(z,c)}.
\end{aligned}
\label{eq:opd_decomposition}
\end{equation}

The first component $O$ is the cross-family OPD signal already induced by the low-capability reference $\mathcal{R}$.
It captures the log-likelihood discrepancy between $\mathcal{R}$ and the current student, forming the baseline OPD signal to which the strong teacher $\mathcal{T}$ adds its within-family log-likelihood shift.
The second component $\Delta_T$ measures the log-likelihood change from $\mathcal{R}$ to $\mathcal{T}$ on the same student action.
Since this comparison is entirely within the teacher family, it provides a more direct signal of which student actions gain or lose support as teacher capability increases.

Standard cross-family OPD transfers the combined signal $O+\Delta_T$ to the student.
When the offset and the within-family shift favor opposite directions, an offset with greater magnitude causes the update to follow the reference-induced direction.
Figure~\ref{fig:motivation}(c) shows that the strong teacher $\mathcal{T}$ and the low-capability reference $\mathcal{R}$ produce the same OPD update direction at a substantial fraction of aligned units.
At these positions, the teacher-family shift affects the update magnitude without reversing the direction already induced by the reference.

This observation motivates CompassOPD.
We remove the shared cross-family offset $O$ and use $\Delta_T$ as the capability-sensitive distillation signal.
The goal remains to transfer the strong teacher model's capability, and the within-family likelihood change provides a clearer direction for achieving this goal.
\section{CompassOPD}
\label{sec:method}
In this section, we introduce CompassOPD. Building on the preceding signal decomposition, CompassOPD removes the shared cross-family offset and uses the within-family log-likelihood shift to supervise student-generated trajectories. A frozen student reference policy further anchors the update according to the student's displacement from its initial policy. Together, these teacher-side and student-side comparisons guide capability transfer across model families. The overall workflow is illustrated in Figure~\ref{fig:compassopd-method}.

\subsection{A Capability Signal from the Teacher Family}
Let $\mathcal{T}$ denote the strong teacher and $\mathcal{R}$ a
low-capability reference model from the same teacher family.
Given a prompt $x$, the current student policy $\pi_\theta$ samples
a response, and both frozen models score the same aligned textual
units in that response. For an aligned unit $z$ under context $c$,
CompassOPD removes the shared cross-family offset $O$ from the
strong-teacher OPD signal $A_{\mathcal{T}}$:
\begin{equation}
\begin{aligned}
\Delta_T(z,c)
&= A_{\mathcal{T}}(z,c)-O(z,c) \\
&= (\ell_{\mathcal{T}}-\ell_{\pi_\theta})
   -(\ell_{\mathcal{R}}-\ell_{\pi_\theta}) \\
&= \ell_{\mathcal{T}}(z\mid c)
   -\ell_{\mathcal{R}}(z\mid c).
\end{aligned}
\label{eq:compass_teacher_shift}
\end{equation}
Here, the student likelihood cancels in the subtraction, leaving a
log-likelihood change measured entirely within the teacher family.
A positive $\Delta_T$ means that $\mathcal{T}$ assigns higher likelihood
to the same aligned action than $\mathcal{R}$, while a negative value
means that $\mathcal{T}$ assigns lower likelihood than $\mathcal{R}$. Comparing $\Delta_T$ across candidate continuations under the same
context reveals how their relative preference changes from
$\mathcal{R}$ to $\mathcal{T}$.

\subsection{Anchoring the Student Update}
The teacher-family shift indicates which sampled actions should receive
more or less support, but does not account for changes the student has
already made. We therefore retain a frozen student reference policy
$\pi_{\mathrm{ref}}$, initialized from the student before
OPD training, to anchor updates relative to that starting point.
For each sampled student token $y_t$ under context $c_t$, we measure
the change in its log-likelihood relative to the reference:
\begin{equation}
\Delta_S(y_t,c_t)
=
\log\pi_\theta(y_t\mid c_t)
-
\log\pi_{\mathrm{ref}}(y_t\mid c_t).
\label{eq:compass_student_shift}
\end{equation}

Let $\Delta_{T,t}$ denote the teacher-family shift assigned to student
token position $t$ through text-space alignment.
CompassOPD combines this signal with the student's displacement
to construct the sampled advantage:
\begin{equation}
A^{\mathrm{Compass}}_t
=
\operatorname{sg}
\left[
\Delta_{T,t}
-
\alpha\Delta_S(y_t,c_t)
\right],
\label{eq:compass_advantage}
\end{equation}
where $\operatorname{sg}$ denotes stop-gradient and $\alpha$ controls
the strength of the student reference anchor.

At the beginning of training, $\pi_\theta=\pi_{\mathrm{ref}}$ and
$\Delta_S=0$, so the update is determined entirely by the teacher-family
shift. As the student moves an action's log-likelihood in the direction
favored by the teacher signal, the anchor term $-\alpha\Delta_S$
opposes further movement in that direction.
To interpret this feedback, consider a policy $\pi$ over aligned
textual actions $z$ at a fixed context $c$. The following
reference-regularized objective captures the balance between following
the teacher-family shift and remaining close to the student reference:
\begin{equation}
\begin{aligned}
\mathcal{J}_{\mathrm{Compass}}(\pi;c)
&=
\mathbb{E}_{z\sim\pi(\cdot\mid c)}
\left[
\Delta_T(z,c)
\right] \\
&\quad-
\alpha
D_{\mathrm{KL}}
\left(
\pi(\cdot\mid c)
\Vert
\pi_{\mathrm{ref}}(\cdot\mid c)
\right).
\end{aligned}
\label{eq:compass_regularized_objective}
\end{equation}
To see how the teacher-family shift and the student reference jointly
shape the target distribution, we examine the maximizer of this objective. For $\alpha>0$, it takes the form
(see Appendix~\ref{app:compass_optimal_policy} for a proof):
\begin{equation}
\pi^*(z\mid c)
=
\frac{
\pi_{\mathrm{ref}}(z\mid c)
\exp\left(\Delta_T(z,c)/\alpha\right)
}{
Z(c)
},
\label{eq:compass_target_policy}
\end{equation}
where $Z(c)$ is the normalization factor ensuring that
$\sum_z \pi^*(z\mid c)=1$. The exponential factor reweights the student's reference policy
according to the teacher-family shift, expressing the transferred
changes relative to the student's own initialization.
In training, we substitute $A^{\mathrm{Compass}}_t$ for the standard
OPD advantage and otherwise retain the same on-policy optimization
procedure.
\begin{table*}[t]
\centering
\setlength{\tabcolsep}{2.2pt}
\scalebox{0.98}{
\begin{tabular}{lcccccccc}
\toprule
Method
& AIME24 & AIME25 & AIME26
& HMMT25$_\text{Feb}$ & HMMT25$_\text{Nov}$ & HMMT26
& MATH-500 & Avg. \\
\midrule

\multicolumn{9}{c}{\textit{Student: Granite4.1-3B}} \\
\ Base
& 4.79 & 7.92 & 5.83 & 1.25 & 1.46 & 3.03 & 64.31 & 12.66 \\
\  SFT
& 21.25 & 19.58 & 19.79 & 8.54 & 6.67 & 14.39 & 80.52 & 24.39 \\
\  OPD
& 25.00 & 18.96 & 17.29 & \textbf{12.29} & 11.04 & 13.83 & 79.12 & 25.36 \\
\  \textbf{CompassOPD}
& \textbf{33.75} & \textbf{24.58} & \textbf{25.62}
& \textbf{12.29} & \textbf{14.37} & \textbf{21.02}
& \textbf{84.36} & \textbf{30.86} \\
\midrule

\multicolumn{9}{c}{\textit{Student: Qwen3-4B}} \\
\  Base
& 24.79 & 18.33 & 18.12 & 12.50 & 8.33 & 17.80 & 83.25 & 26.16 \\
\  SFT
& 28.33 & 24.17 & 22.92 & 13.54 & 12.50 & 15.53 & 84.51 & 28.79 \\
\  OPD
& 36.25 & 32.29 & 30.83 & 17.50 & 20.00 & 21.02 & 86.70 & 34.94 \\
\  \textbf{CompassOPD}
& \textbf{40.00} & \textbf{32.50} & \textbf{35.42}
& \textbf{18.12} & \textbf{20.21} & \textbf{24.43}
& \textbf{88.85} & \textbf{37.08} \\
\midrule

\multicolumn{9}{c}{\textit{Student: OLMo-3-7B}} \\
\  Base
& 5.63 & 5.21 & 6.88 & 1.67 & 3.12 & 2.65 & 64.51 & 12.81 \\
\  SFT
& 30.00 & 25.83 & 28.54 & 13.33 & 14.17 & 16.67 & 83.95 & 30.36 \\
\  OPD
& 31.46 & 29.17 & 28.54 & 16.88 & 15.00 & 23.30 & 86.14 & 32.93 \\
\  \textbf{CompassOPD}
& \textbf{36.67} & \textbf{30.00} & \textbf{29.38}
& \textbf{17.50} & \textbf{19.79} & \textbf{23.48}
& \textbf{86.38} & \textbf{34.74} \\
\bottomrule
\end{tabular}
}
\caption{
Main results with Qwen3.5-35B-A3B as the teacher and
Qwen3.5-0.8B as the reference for CompassOPD.
OPD and CompassOPD share the same SFT initialization for each student.
Best results within each student group are shown in bold.
}
\label{tab:main-qwen35}
\end{table*}
\begin{table*}[t]
\centering
\setlength{\tabcolsep}{2.2pt}
\scalebox{0.98}{
\begin{tabular}{lcccccccc}
\toprule
Method
& AIME24 & AIME25 & AIME26
& HMMT25$_\text{Feb}$ & HMMT25$_\text{Nov}$ & HMMT26
& MATH-500 & Avg. \\
  \midrule

  Base
  & 4.79 & 7.92 & 5.83
  & 1.25 & 1.46 & 3.03
  & 64.31 & 12.66 \\

  \midrule
  \multicolumn{9}{c}{\textit{Teacher: Qwen3-30B-A3B (Ref. Qwen3-0.6B)}} \\
  SFT
  & 21.04 & 20.62 & 17.71
  & 11.87 & 8.33 & 13.83
  & 79.60 & 24.71 \\
  OPD
  & 22.50 & 25.21 & 23.13
  & 13.33 & 14.37 & 19.13
  & 83.20 & 28.70 \\
  \textbf{CompassOPD}
  & \textbf{28.54} & \textbf{30.21} & \textbf{26.46}
  & \textbf{16.88} & \textbf{15.00} & \textbf{21.97}
  & \textbf{84.45} & \textbf{31.93} \\

  \midrule
  \multicolumn{9}{c}{\textit{Teacher: Ministral-3-14B (Ref. Ministral-3-3B)}} \\
  SFT
  & 18.54 & 18.12 & 14.79
  & 9.79 & 6.25 & 13.26
  & 78.44 & 22.74 \\
  OPD
  & 21.46 & 16.67 & 18.12
  & \textbf{11.46} & 8.75 & 13.26
  & 78.93 & 24.09 \\
  \textbf{CompassOPD}
  & \textbf{22.29} & \textbf{24.17} & \textbf{21.46}
  & \textbf{11.46} & \textbf{10.42} & \textbf{16.67}
  & \textbf{80.01} & \textbf{26.64} \\

  \bottomrule
  \end{tabular}
  }
  \caption{
  Results with Qwen3 and Mistral teachers on Granite4.1-3B.
Within each teacher group, OPD and CompassOPD share the same
teacher-specific SFT initialization.
Best results are shown in bold.
  }
  \label{tab:teacher-family}
  \end{table*}
\paragraph{MoE self-reference.}
To obtain a teacher-family reference without a separate checkpoint,
we also consider using an MoE teacher under two expert activation
configurations (Figure~\ref{fig:compassopd-method}(c)).
The standard configuration serves as $\mathcal{T}$, while a
configuration with fewer activated experts serves as $\mathcal{R}$,
sharing the same frozen weights.
Their log-likelihood difference on the same aligned student-generated
actions supplies $\Delta_T$ in
Equation~\ref{eq:compass_teacher_shift}.
\section{Experiments}
\label{sec:experiments}
\subsection{Experimental Setup}
\paragraph{Training Data and Benchmarks.}
We evaluate CompassOPD on reasoning tasks. We use DAPO-Math~\cite{yu2025dapoopensourcellmreinforcement} and DeepScaleR-Preview~\cite{deepscaler2025} for on-policy training. Evaluation is conducted on seven benchmarks: AIME 2024~\cite{aime24}, AIME 2025~\cite{aime25}, AIME 2026~\cite{aime26}, HMMT 2025 February, HMMT 2025 November, HMMT 2026~\cite{dekoninck2026matharena}, and MATH-500~\cite{lightman2023letsverifystepstep}. For each benchmark, we report mean accuracy over 16 independent generations per problem.
\paragraph{Models and baselines.}
We evaluate CompassOPD across three teacher families:
Qwen3.5, Qwen3, and Mistral. The teachers are Qwen3.5-35B-A3B in non-thinking mode,
Qwen3-30B-A3B-Instruct-2507, and Ministral-3-14B~\cite{liu2026ministral3}, respectively.
Each teacher is paired with a lower-capability reference from
its own family.
Our student models span three model families, represented by
Granite4.1-3B~\cite{ibm2026granite41}, Qwen3-4B, and OLMo-3-7B~\cite{olmo2026olmo3}. We compare the original model (\textbf{Base}), the teacher-trajectory
\textbf{SFT} checkpoint, standard sampled \textbf{OPD}, and
\textbf{CompassOPD}.
For each comparison, both distillation methods start from the same
SFT checkpoint trained on reasoning trajectories generated by the
strong teacher.
Further analyses examine teacher scale within Qwen3.5 with a fixed
0.8B reference and MoE self-reference constructed by reducing
expert activation (see Appendix~\ref{app:implementation} for details).
\paragraph{Implementation details.}
We implement all methods using VERL~\cite{sheng2024hybridflow} and use vLLM~\cite{kwon2023efficientmemorymanagementlarge} for sampling generation and teacher scoring. The learning rate is set to $1\times10^{-6}$, with a batch size of 128. The maximum prompt length and response length are 2,048 and 24,576 tokens, respectively. For CompassOPD, we set $\alpha=0.5$. OPD and CompassOPD use the same student initialization, token/span alignment.

\subsection{Main Results}
Tables~\ref{tab:main-qwen35} and~\ref{tab:teacher-family} report cross-family distillation results spanning
three teacher families and three student families.
CompassOPD consistently outperforms standard OPD
across all five configurations.
\paragraph{Performance across student families.}
Table~\ref{tab:main-qwen35} compares the three student families with
Qwen3.5-35B-A3B as the teacher.
CompassOPD improves upon OPD in 20 of the 21 student--benchmark
combinations, with one tie.
The gains are most pronounced on Granite4.1-3B, where standard OPD
provides limited improvement beyond SFT.
CompassOPD also improves Qwen3-4B and OLMo-3-7B, where standard OPD
already yields clear gains, demonstrating benefits across students
with different responses to cross-family distillation.

\paragraph{Performance across teacher families.}
Table~\ref{tab:teacher-family} compares OPD and CompassOPD using Qwen3 and Mistral teachers
with Granite4.1-3B as the student.
CompassOPD improves average performance under both teacher families,
outperforming OPD in 13 of the 14 teacher--benchmark combinations,
with one tie.
Together with the Qwen3.5 results, these findings demonstrate
consistent improvements across all three evaluated teacher families.
\begin{figure}[t]
\centering

\subfloat[Anchor sensitivity.]{
    \includegraphics[width=0.48\columnwidth]
    {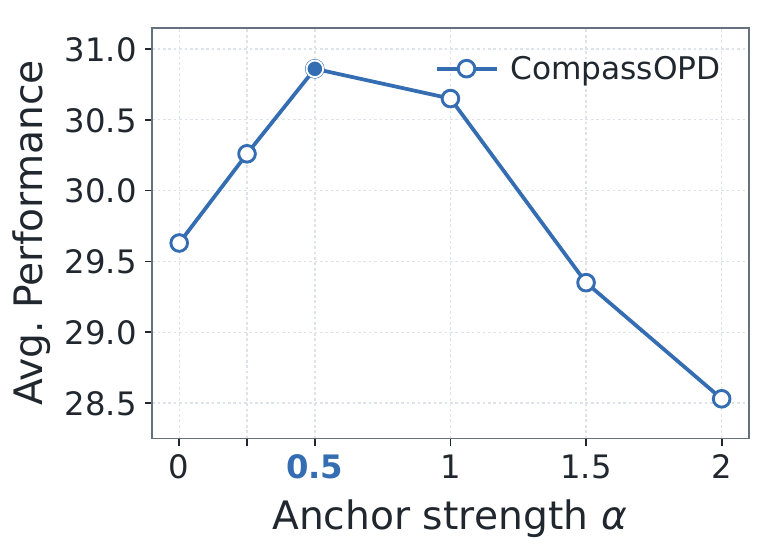}
    \label{fig:analysis_anchor}
}
\subfloat[Offset restoration.]{
    \includegraphics[width=0.48\columnwidth]
    {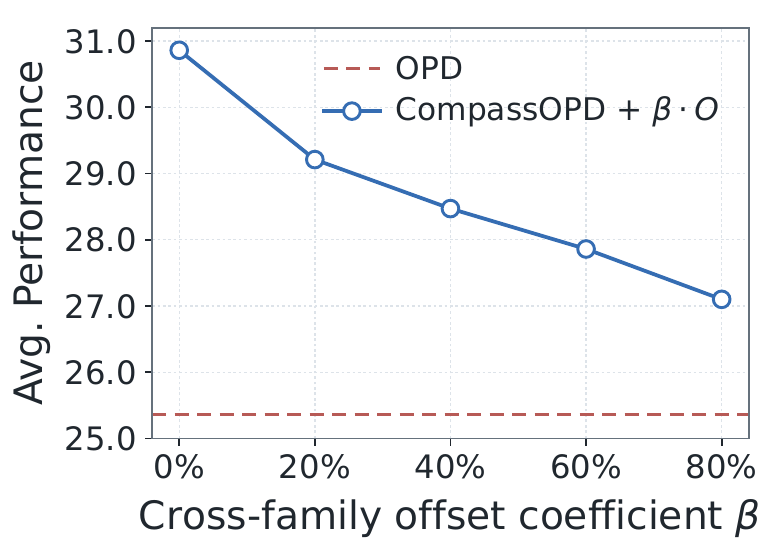}
    \label{fig:analysis_offset}
}

\caption{
Component analysis of CompassOPD on Granite4.1-3B, with
Qwen3.5-35B-A3B as the teacher and Qwen3.5-0.8B as the reference.
(a) Student anchor sensitivity.
(b) Offset restoration with $\alpha=0.5$;
the dashed line denotes standard OPD.
}
\label{fig:component_analysis}
\end{figure}
\subsection{Component Analysis}
\paragraph{Sensitivity to the student reference anchor.}
Figure~\ref{fig:component_analysis}(a) shows that performance
improves as the anchor strength increases from $\alpha=0$ to
$\alpha=0.5$, where it reaches its highest value.
When $\alpha=0$, the update relies entirely on the teacher-family
shift; a moderate anchor additionally accounts for the student's
displacement from its initial policy.
Performance remains competitive at $\alpha=1.0$ but declines under
stronger anchoring, suggesting that excessive regularization
restricts student updates.
We set $\alpha=0.5$ in our main experiments.

\paragraph{Impact of the offset on cross-family OPD.}
To test whether the cross-family offset identified in
Section~\ref{sec:pilot} limits capability transfer, we gradually
restore it while keeping the student anchor fixed at $\alpha=0.5$:
\begin{equation}
A_t(\beta)
=
\operatorname{sg}\left[
\Delta_{T,t}+\beta O_t-\alpha\Delta_S(y_t,c_t)
\right].
\end{equation}
Here, $O_t$ is the offset assigned to student token position $t$,
and $\beta$ controls the fraction restored, with $\beta=0$
recovering CompassOPD.
Figure~\ref{fig:component_analysis}(b) shows that the average
accuracy of Granite4.1-3B decreases monotonically as more of the
offset is restored, removing most of the gain over OPD.
This decline under a fixed student anchor demonstrates the benefit
of removing the cross-family offset for capability transfer.

\begin{table*}[t]
\centering
\setlength{\tabcolsep}{2.15pt}
\scalebox{0.96}{
\begin{tabular}{lcccccccc}
\toprule
Configuration
& AIME24 & AIME25 & AIME26
& HMMT25$_{\text{Feb}}$
& HMMT25$_{\text{Nov}}$
& HMMT26
& MATH-500
& Avg. \\
\midrule

OPD
& 25.00 & 18.96 & 17.29
& 12.29 & 11.04 & 13.83
& 79.12 & 25.36 \\

\midrule
\multicolumn{9}{l}{
\text{CompassOPD with $\mathcal{R}=0.8\mathrm{B}$}
} \\
\quad $\mathcal{T}=35\mathrm{B}\text{-A}3\mathrm{B}$
& 33.75 & 24.58 & 25.62
& 12.29 & 14.37 & 21.02
& 84.36 & \textbf{30.86} \\
\quad $\mathcal{T}=9\mathrm{B}$
& 31.25 & 22.92 & 23.75
& 12.92 & 13.33 & 18.56
& 83.76 & 29.50 \\
\quad $\mathcal{T}=4\mathrm{B}$
& 30.21 & 23.13 & 24.58
& 12.08 & 9.17 & 17.80
& 83.61 & 28.65 \\
\quad $\mathcal{T}=2\mathrm{B}$
& 24.37 & 19.79 & 21.67
& 12.92 & 9.58 & 15.53
& 81.66 & 26.50 \\

\bottomrule
\end{tabular}
}
\caption{
Results on Granite4.1-3B with Qwen3.5 teachers of different
scales. CompassOPD uses a fixed Qwen3.5-0.8B reference,
while OPD uses the 35B-A3B teacher.
}
\label{tab:teacher-scale}
\end{table*}
\subsection{Teacher and Reference Configurations}
\paragraph{Effect of teacher--reference scale separation.}
We examine how the scale separation between $\mathcal{T}$ and
$\mathcal{R}$ affects relative supervision.
With Qwen3.5-0.8B fixed as $\mathcal{R}$, we progressively reduce
$\mathcal{T}$ from Qwen3.5-35B-A3B to 9B, 4B, and 2B.
Table~\ref{tab:teacher-scale} shows that the average accuracy of
Granite4.1-3B decreases as the teacher approaches the reference
in scale, consistent with the within-family shift becoming less
informative as their separation narrows.
Nevertheless, the 2B teacher paired with the 0.8B reference still
outperforms standard OPD with the 35B-A3B teacher.
Thus, even modest within-family variation can provide more effective
supervision than the absolute likelihoods of a much larger
cross-family teacher model.
This reliance on a teacher-family reference raises a practical
question: what if the teacher family lacks a suitable low-capability
checkpoint to serve as $\mathcal{R}$?
\begin{table}[t]
\centering
\small
\setlength{\tabcolsep}{6pt}
\begin{tabular}{llc}
\toprule
Method & Reference $\mathcal{R}$ & Avg. \\
\midrule
SFT & -- & 24.39 \\
OPD & -- & 25.36 \\
\midrule
CompassOPD & Separate checkpoint (0.8B) & 30.86 \\
CompassOPD & Teacher self-reference & 28.79 \\
\bottomrule
\end{tabular}
\caption{
Reference constructions for Granite4.1-3B with Qwen3.5-35B-A3B
as the teacher. Self-reference uses the same teacher checkpoint
with expert activation reduced to approximately 2B active parameters.
}
\label{tab:self-reference}
\end{table}
\paragraph{Can a teacher model provide its own reference?}
CompassOPD relies on a suitable low-capability reference within
the teacher family to remove the offset.
A practical concern is that some top-tier very large models may
not offer publicly accessible smaller versions within the same
family, such as Kimi K3~\cite{kimiteam2026kimik3openfrontier}.
However, these models commonly use MoE architectures, which provide
a natural alternative: the same checkpoint can be run under
different activation capacities.
We therefore use Qwen3.5-35B-A3B under its standard routing
configuration as $\mathcal{T}$, and construct a reference
$\mathcal{R}$ with roughly 2B activated parameters by enabling
only one shared expert and one fine-grained expert.
This design varies the activation capacity while preserving the
checkpoint and model family, allowing the teacher model to provide
its own within-family reference.

Table~\ref{tab:self-reference} shows that this self-reference
configuration improves average accuracy over standard OPD by
3.43 points.
Although it trails the dedicated 0.8B reference by 2.07 points,
it retains a substantial improvement without requiring a separately
released reference checkpoint.
The result shows that useful within-family variation can come from
different activation capacities of a single teacher, providing a
practical alternative when a suitable reference model is unavailable.
\section{Related Work}
\paragraph{Cross-family distillation.}
Different training recipes and tokenizers complicate distribution matching in cross-family
distillation.
Sequence-level distillation transfers knowledge through teacher-generated
text, which students can process with their own
tokenizers~\cite{kim-rush-2016-sequence}.
For finer-grained supervision,
DSKD~\cite{zhang2024dualspaceknowledgedistillationlarge} combines
dual-space distillation with cross-model attention to align
representations and distributions.
ULD~\cite{boizard2025crosstokenizerdistillationuniversallogit}
uses optimal transport to compare distributions across vocabularies,
while ALM~\cite{minixhofer2025universalcrosstokenizerdistillationapproximate}
matches likelihoods on corresponding text segments.
Recent work extends cross-tokenizer distillation to student-generated
trajectories through shared text units, token mapping, or byte-prefix
marginalization~\cite{sun2026simctrecoveringlostsupervision,niu2026breakingtokenizerbarrieronpolicy,wang2026crosstokenizeronpolicydistillationbyteprefix}.
Together, these methods provide comparable supervision interfaces
across model families.

\paragraph{On-policy distillation.}
On-policy distillation obtains teacher feedback on student-generated
trajectories, covering the states visited by the student.
GKD~\cite{agarwal2024onpolicydistillationlanguagemodels} studies
distillation on student-generated sequences with different divergence
objectives, while MiniLLM~\cite{gu2026minillmonpolicydistillationlarge}
adopts reverse KL optimization.
We follow the sampled OPD formulation of
\citet{lu2025onpolicydistillation}, using teacher--student
log-probability differences as token-level advantages.
Recent analysis shows that transfer also depends on reasoning-pattern
compatibility and the teacher's additional capabilities, beyond its
standalone performance~\cite{li2026rethinkingonpolicydistillationlarge}.
Concurrently with our work, other studies have explored contrastive
supervision in OPD. RLCSD~\cite{pan2026rlcsdreinforcementlearningcontrastive}
contrasts correct and incorrect solution hints to mitigate style drift,
while W2S-OPD~\cite{yu2026weaktostrongonpolicydistillation}
aims to improve a stronger student by leveraging differences between smaller models within the same family, thereby enabling weak-to-strong OPD. These methods pursue objectives distinct
from ours. 
CompassOPD is the first work to identify the cross-family offset as
a central limitation in cross-family OPD and to propose removing it while
using within-family likelihood shifts for cross-family capability transfer.
\section{Conclusion}
This paper reveals a key issue in cross-family OPD: standard OPD mixes the cross-family offset with within-family likelihood changes, allowing the former to mask capability-sensitive supervision. CompassOPD removes this offset, performs distillation using relative likelihood changes within the teacher family, and anchors student updates to the student's initial policy. Experiments covering three student families and multiple teacher families show that CompassOPD consistently outperforms standard cross-family OPD. Further analysis validates the roles of offset removal and student-side anchoring. The self-reference construction based on MoE expert activation also demonstrates that the method can be applied without an independent small-scale reference checkpoint. Overall, relative changes within the teacher family provide a more effective supervision interface for cross-family capability transfer.
\section*{Limitations}
Due to computational constraints, we have not evaluated CompassOPD
on very large models.
Nevertheless, our experiments across multiple teacher and student
sizes demonstrate its effectiveness at different model scales.
Extending this validation to substantially larger models remains
a direction for future work.



\bibliography{custom}

\appendix
\newpage
\begin{center}
    \Large \textbf{Appendix}
\end{center}
\section{Implementation Details}
\label{app:implementation}
\subsection{Model Configuration}
The main experiments use Qwen3.5-35B-A3B as the strong teacher
$\mathcal{T}$ and Qwen3.5-0.8B as the teacher-family reference
$\mathcal{R}$, with both models running in non-thinking mode.
The student models are Granite4.1-3B, Qwen3-4B, and
OLMo-3-7B-Instruct-SFT.

For teacher-family generalization, we use
Qwen3-30B-A3B-Instruct-2507 with Qwen3-0.6B, and
Ministral-3-14B-Instruct-2512 with Ministral-3-3B-Instruct-2512,
as teacher--reference pairs for Granite4.1-3B.
For the teacher––reference scale separation experiments, we fix Qwen3.5-0.8B as
$\mathcal{R}$ and vary $\mathcal{T}$ across Qwen3.5-35B-A3B,
9B, 4B, and 2B.
All four configurations use the same Granite4.1-3B SFT checkpoint
trained on trajectories generated by Qwen3.5-35B-A3B.

During distillation, the strong teacher $\mathcal{T}$,
the teacher-family reference $\mathcal{R}$, and the frozen copy
of the student's initial policy $\pi_{\mathrm{ref}}$ remain fixed.
Only the current student policy $\pi_\theta$ is updated.
Standard OPD and CompassOPD start from the same SFT checkpoint
in every comparison.

\subsection{Cold Start Settings}
Cross-family models often differ in response style, format,
and reasoning length.
Following prior work on vocabulary alignment~\cite{sun2026simctrecoveringlostsupervision},
we first perform supervised fine-tuning on the student using
reasoning trajectories generated by the strong teacher to reduce
these differences before on-policy distillation.
The resulting checkpoint initializes both standard OPD and CompassOPD.

We use the OpenThoughts-114k~\cite{guha2025openthoughtsdatarecipesreasoning} dataset, which
includes 89,120 math problems, 19,904 code problems, and
4,933 problems from biology, physics, chemistry, and logic puzzles.
For each teacher family, we replace the original dataset answers
with responses generated by its strong teacher in non-thinking mode.
We retain only samples that terminate normally and have non-empty
responses.

The cold-start stage uses full-parameter supervised fine-tuning
with a maximum sequence length of 32,768 tokens.
We use a learning rate of $2\times10^{-5}$, weight decay of
$10^{-6}$, a cosine learning rate schedule with 10\% warmup,
and a batch size of 32, and train for one epoch.
This stage provides a common initialization adapted to the
teacher's responses.
For each teacher--student combination, standard OPD and CompassOPD
use the same cold-start data and SFT checkpoint.

\subsection{Distillation Settings}
Based on model capability, we use the DAPO-Math dataset for Qwen3-4B,
and the DeepScaleR-Preview dataset for
Granite4.1-3B and OLMo-3-7B.
Rollouts use temperature 1.0 and top-$p=1.0$, with maximum prompt
and response lengths of 2,048 and 24,576 tokens, respectively.

The optimizer is AdamW with a constant learning rate of
$1\times10^{-6}$ and no warmup.
The PPO mini-batch size is 64, and the micro-batch size per GPU is 1.
The training objective uses the distillation signal without
task correctness rewards. Within each comparison, standard OPD and CompassOPD use the
same number of training steps and sampled student responses.
All experiments are conducted on eight NVIDIA H200 GPUs.

\subsection{Cross-Tokenizer Alignment}
For each student-generated response, we score the same text
with the strong teacher and the teacher-family reference.
Token sequences across models are monotonically aligned according
to shared text boundaries.
For one-to-one token matches, we use the corresponding token
log-likelihood directly.
For text segments represented by different numbers of tokens,
we construct a common span and average the token log-likelihoods
within that span to reduce scale differences associated with
tokenizer granularity.

For CompassOPD, the teacher-family shift is computed from the
aligned teacher and reference scores and assigned to every student
token within the corresponding span.
The student-reference displacement is computed directly at each
student token position.
Positions where consistent text boundaries cannot be formed are
excluded by an alignment mask and do not contribute to the loss.
Standard cross-family OPD and CompassOPD use the same alignment
procedure.

\subsection{MoE Self-Reference}
In the self-reference experiments, $\mathcal{T}$ and $\mathcal{R}$
share the same Qwen3.5-35B-A3B checkpoint.
The strong teacher uses the default MoE routing configuration,
activating eight fine-grained experts per token in addition to
the shared expert.
The reference configuration activates one fine-grained expert
while retaining the shared expert, corresponding to approximately
2B activated parameters.
The two configurations share all model weights, the tokenizer,
and the input context, and differ only in expert activation.
We perform two separate forward passes with frozen weights and
use the difference between their aligned log-likelihoods as
$\Delta_T$.

\subsection{Evaluation Protocol}
All models use their native chat templates.
During evaluation, the maximum generation length is 24,576 tokens,
the temperature is 0.7, top-$p=1.0$, top-$k=40$, and the presence
penalty is 2.0.

Each model independently samples 16 responses per problem on benchmarks.
For each benchmark, we report accuracy averaged over all sampled
responses.
The Avg.\ column reports the unweighted mean of the seven
benchmark accuracies.
Evaluation uses the automatic scoring pipeline of EvalScope~\cite{evalscope_2024}.

\section{Analysis of the CompassOPD Target Policy}
\label{app:compass_optimal_policy}

We analyze Equation~\ref{eq:compass_regularized_objective} at a fixed
context $c$, holding $\Delta_T$ and $\pi_{\mathrm{ref}}$ fixed.
All policies are defined over the same aligned textual action space.
Assume $\alpha>0$, that $\pi_{\mathrm{ref}}$ is positive on this
action space, and that the quantities below are finite.
Define the normalization factor
\begin{equation}
Z(c)
=
\sum_z
\pi_{\mathrm{ref}}(z\mid c)
\exp\left(\Delta_T(z,c)/\alpha\right).
\end{equation}

The policy in Equation~\ref{eq:compass_target_policy} satisfies
\[
\log
\frac{\pi^*(z\mid c)}
     {\pi_{\mathrm{ref}}(z\mid c)}
=
\frac{\Delta_T(z,c)}{\alpha}
-
\log Z(c).
\]
Substituting this identity into
Equation~\ref{eq:compass_regularized_objective} yields
\begin{equation}
\begin{aligned}
\mathcal{J}_{\mathrm{Compass}}(\pi;c)
&=
\alpha\log Z(c) \\
&\quad-
\alpha D_{\mathrm{KL}}
\left(
\pi(\cdot\mid c)
\Vert
\pi^*(\cdot\mid c)
\right).
\end{aligned}
\end{equation}
Since $Z(c)$ is independent of $\pi$ and the KL divergence is
nonnegative, the objective is bounded above by $\alpha\log Z(c)$,
with equality if and only if
$\pi(\cdot\mid c)=\pi^*(\cdot\mid c)$.
This establishes the optimality of
Equation~\ref{eq:compass_target_policy}.

\end{document}